\documentclass[preprintreview12pt]{elsarticle}

\usepackage{multirow}
\usepackage{graphicx}
\usepackage{longtable}
\usepackage[table]{xcolor}% http://ctan.org/pkg/xcolor
\usepackage{listings}
\usepackage{multicol}
\usepackage{hyperref}

\usepackage{graphicx}
\usepackage{amssymb}
\usepackage{amsmath}
 
\journal{International Journal of Forecasting}

\begin{document}

\begin{frontmatter}

\title{Blending gradient boosted trees and neural networks for point and probabilistic forecasting of hierarchical time series}

\author{Ioannis Nasios}
\author{Konstantinos Vogklis}
\address{Nodalpoint Systems, Athens, Greece}

\begin{abstract}

In this paper we tackle the problem of point and probabilistic forecasting by describing a blending methodology of machine learning models that belong to gradient boosted trees and neural networks families. These principles were successfully applied in the recent M5 Competition on both Accuracy and Uncertainty tracks.  The keypoints of our methodology are: a) transform the task to regression on sales for a single day b) information rich feature engineering c) create a diverse set of state-of-the-art machine learning models and d) carefully construct validation sets for model tuning. We argue that the diversity of the machine learning models along with the careful selection of validation examples, where the most important ingredients for the effectiveness of our approach. Although forecasting data had an inherent hierarchy structure (12 levels), none of our proposed solutions exploited that hierarchical scheme. Using the proposed methodology, our team was ranked within the gold medal range in both Accuracy and the Uncertainty track. Inference code along with already trained models are available at \url{https://github.com/IoannisNasios/M5_Uncertainty_3rd_place}
%and won the 3rd prize in the  Uncertainty track.
\end{abstract}


\end{frontmatter}

\section{Introduction}
\label{sec:introduction}

Machine Learning (ML) methods have been well established in the academic literature as alternatives to statistical ones for time series forecasting~\cite{makridakis2022m5, MAKRIDAKIS202054, makridakis2018statistical, kim2015box,CRONE2011635}. The results of the recent M-competitions also reveal such a trend from the classical Exponential Smoothing and ARIMA methods~\cite{hyndman2018forecasting} towards more data-driven generic ML models such as Neural Networks~\cite{bishop1995neural} and Gradient Boosted Trees~\cite{friedman2001greedy, friedman2002stochastic}.

%\subsection{Machine Learning in Forecasting}
One of the most important results of the recent M5 Competition~\cite{makridakis2022m5} was the superiority of Machine Learning methods, particularly the Gradient Boosted Trees and Neural Networks, against statistical time series methods (e.g., exponential smoothing, ARIMA etc.). This realization is in the opposite direction of the findings of previous M competitions that classical time series methods were more accurate, and establishes a milestone in the domination of Machine Learning in yet another scientific domain.

The M5 forecasting competition was designed to empirically evaluate the accuracy of new and existing forecasting algorithms in a real-world scenario of hierarchical unit sales series.
The dataset provided contains $42,840$ hierarchical sales data from Walmart. It covers stores in three US states (California Texas and Wisconsin) and includes sales data per item level, department and product categories and store details ranging from February 2011 to April 2016. The products have a (maximum) selling history of $1941$ days. 
The competition was divided into two tracks, one requiring point forecasts (Accuracy track), and one requiring the estimation of the uncertainty distribution (Uncertainty track).
The time horizon for both tracks was 28 days ahead. For the Accuracy track, the task was to predict the sales for each one of the  $42,840$ hierarchical time series following day $1941$. For the Uncertainty track, the task was to provide probabilistic forecasts for the corresponding median and four prediction intervals ($50\%$, $67\%$, $95\%$, and $99\%$).

Table~\ref{tab:series} presents all hierarchical groupings of the data. Level~12, containing $30,490$ unique combinations of product per store, is the most disaggregated level. Following the competition rules, only these $30,490$ series needed to be submitted; the forecasts of all higher levels would be automatically calculated by aggregating (summing) the ones of this lowest level. So, regardless the way we used the hierarchy information to produce 28-day ahead predictions, all we needed to actually submit was level~12 series only.

\begin{table}[!h]
\begin{footnotesize}
\begin{tabular}{lp{6.5cm}p{2cm}r}
\hline
  Level & Level Description                                                & Aggr. Level & \multicolumn{1}{r|}{\#of series} \\ \hline
1  & Unit sales of all products, aggregated for all stores/states           & Total               & 1                                       \\
2  & Unit sales of all products, aggregated for each State                  & State               & 3                                       \\
3  & Unit sales of all products, aggregated for each store                  & Store               & 10                                      \\
4  & Unit sales of all products, aggregated for each category               & Category            & 3                                       \\
5  & Unit sales of all products, aggregated for each department             & Department          & 7                                       \\
6  & Unit sales of all products, aggregated for each State and   category   & State/Category      & 9                                       \\
7  & Unit sales of all products, aggregated for each State and department   & State/Department    & 21                                      \\
8  & Unit sales of all products, aggregated for each store and   category   & Store/Category      & 30                                      \\
9  & Unit sales of all products, aggregated for each store and   department & Store/Department    & 70                                      \\
10 & Unit sales of product x,   aggregated for all stores/states            & Product             & 3,049                                   \\
11 & Unit sales of product x,   aggregated for each State                   & Product/State       & 9,147                                   \\
12 & Unit sales of product x,   aggregated for each store                   & Product/Store       & 30,490                                  \\
\multicolumn{3}{l}{Total}                                                                         & 42,840                                 
\end{tabular}
\end{footnotesize}
\caption{M5 series aggregation levels}
\label{tab:series}
\end{table}
 
\section{Machine Learning based forecasting}
\label{sec:mlforecast}

Machine Learning methods are designed to learn patterns from data and they make no assumptions about their nature. Time series forecasting can be easily formulated as a supervised learning task. The goal is to approximate a function $f(\cdot; \theta ) : \mathbf{R}^d \rightarrow \mathbf{R}$ controlled by a set of parameters $\theta$, which corresponds to the relation between a vector of input variables (features) and a target quantity.  The machine learning setup is completed once we define: a) a (training) dataset $\mathcal{D}={(x^{(k)}
,y^{(k)})}, \ \  k=1...n $ consisting of a set of tuples containing features $x^{(k)} \in \mathbf{R}^d$ that describe a target $y^{(k)} \in \mathbf{R}$ (number of daily sales in the context of M5 competition) and  b) a suitable loss function to be minimized $\mathcal{L}(Y, f(x;\theta))$.  The parameters of $f(\cdot; \theta)$ are then tuned to minimize the loss function $\mathcal{L}$ using the tuples of the training dataset $\mathcal{D}$.

\subsection{Feature engineering}
\label{sec:engineering}
One of the key ingredients in any Machine Learning method is the creation of representative and  information-rich input features $x^{(\cdot)} \in \mathbf{R}^d$. The task of feature engineering is a largely ad hoc procedure, considered equal parts science and art, and it crucially depends on the experience of the practitioner on similar tasks. In the context of the M5 Competition, we worked with the following feature groups:  
\begin{enumerate}
    \item Categorical id-based features: This is a special type of features that take only discrete values. Their numerical encoding can take many forms and depends on the methodology chosen. 
    \begin{enumerate}
        \item  Categorical variables encoding via mean target value:  It is a process of encoding the target in a predictor variable perfectly suited for  categorical variables. Each category is replaced with the  corresponding probability of the target value in the presence of this category~\cite{potdar2017comparative}.
        \item Trainable embedding encoding: A common paradigm that gained a lot of popularity since its application to Natural Language Processing~\cite{al2013polyglot,akbik2019flair} is to project the distinct states of a categorical feature to a real-valued, low-dimensional latent space.This is usually implemented as a Neural Network layer (called Embedding Layer) and it is used to compactly encode all the discrete states of a categorical feature. 
        Contrary to one hot encoding which are binary, sparse, and very high-dimensional, trainable embeddings are low-dimensional floating-point vectors (see Fig.~\ref{fig:embeddings}):
        \begin{figure}[!h]
            \centering
            \includegraphics[scale=0.25]{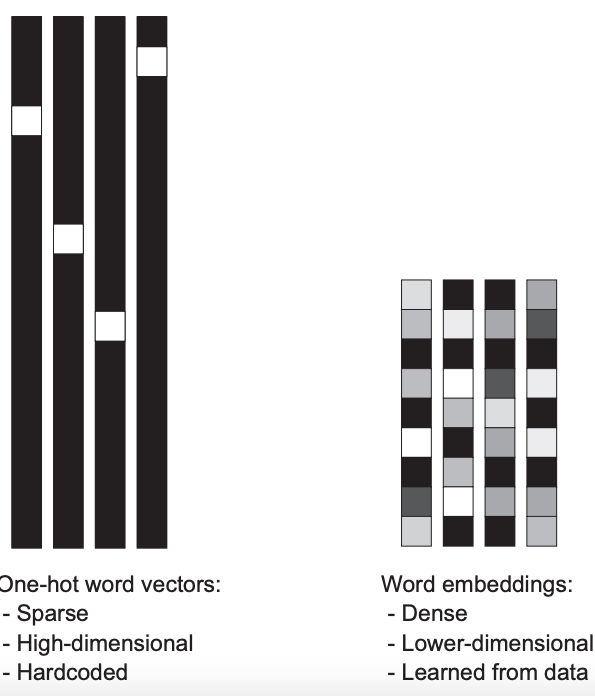}
            \caption{Trainable embeddings vs one hot encoding. Image taken from~\cite{chollet2018deep}}
            \label{fig:embeddings}
        \end{figure}

    \end{enumerate}
    \item Price related: Sell prices, provided on a week level for each combination of store and product.  Prices are constant at weekly basis, although they may change through time. Using this information we calculate statistical features such as maximum, minimum, mean,  standard deviation and also number of unique historical prices for each combination of store and product (level 12).
    \item Calendar related 
    \begin{itemize}
        \item Special events and holidays (e.g. Super Bowl, Valentine’s Day, and Orthodox Easter), organized into four classes, namely Sporting, Cultural, National, and Religious. 
        \item Supplement Nutrition Assistance Program (SNAP)  activities that serve as promotions. This is a binary variable (0 or 1) indicating whether the stores of CA, TX or WI allow SNAP purchases on the examined date.
    \end{itemize}
    \item Lag related features:
    \begin{itemize}
        \item Lag only: These features are based the historical sales for each store/product combination (level 12) for $28+ k$ days before a given date $t$ with $k$ ranging from $1$ to $14$, thus spanning two weeks. 
      
        \item Rolling only: Rolling mean and standard deviation of  historical sales for each store/product (level 12) ending 28 days before a given date $t$. 
  
        \item Lag and rolling:  Rolling mean and standard deviation until a lag date in the past.
    \end{itemize}

\end{enumerate}

In total we devised around $80$ input features, each one used to predict the unit sales of a specific product/store (Level 12) for one specific date.

\subsection{Cross Validation} 
\label{sec:validation}
The inherent  ordering of time series forecasting (i.e. the time component) forces  ML practitioners to define special cross validation schemes and avoid k-fold validation~\cite{bengio2004no} random splits back and forth in time. We chose the following different training/validation splits. 
\begin{itemize}
\item	Validation split 1:
        \begin{itemize}
            \item Training days  $[d_1, d_2, \ldots, d_{1940}] $
            \item Validation days $[d_{1914}, d_{1915}, \ldots d_{1941}]$ (last 28 days)
        \end{itemize}
        
\item	Validation split 2: 
        \begin{itemize}
            \item Training days  $[d_1, d_2, \ldots, d_{1885}]$
            \item Validation days $[d_{1886}, d_{1887}, \ldots,d_{1913}]$ (28 days before last 28 days)
        \end{itemize}

\item 	Validation split 3: 
        \begin{itemize}
            \item Training days  $[d_1, d_2, \ldots, d_{1577}]$ 
            \item Validation days $[d_{1578}, d_{1579}, \ldots, d_{1605}]$ (exactly one year before) 
        \end{itemize}
\end{itemize}

Modeling and blending was based on improving the mean of the competition metric on these splits. For each split we followed a two step modeling procedure:
\begin{enumerate}
    \item[]  {\bf Tuning}: Use all three validation sets and the competition metric to select the best architecture and parameters of the models.   
    \item [] {\bf Full train}: Use the fine-tuned parameters of the previous step to perform a full training run until day $d_{1941}$.
\end{enumerate}

The biggest added value of using multiple validation sets was the elimination of any need for external adjustments on the final prediction(see Finding 5 in ~\cite{makridakis2022m5}).

\section{Point Forecast Methodology}
\label{sec:accuracy}

Point forecast is based on ML models that predict the number of daily sales for a specific date and a specific product/store combination. In order to forecast the complete $28$ days horizon we apply a recursive multi-step scheme~\cite{taieb2012recursive} . This involves using the prediction of the model in a specific time step (day)  as an input in order to predict the subsequent time step. This process is repeated until the desired number of steps have been forecasted. We found this methodology to be superior to a direct multi-step scheme were the models would predict at once $28$ days in the future.

The performance measure selected for this forecasting task was a variant of Mean Absolute Scaled Error (MASE) \cite{hyndman2006another} called  Root Mean Squared Scaled Error (RMSSE). The measure for the 28 day horizon is defined in Eq.~\ref{eqn:rmsse}
\begin{equation}
\displaystyle
RMSSE = \sqrt{\frac{\frac{1}{h} \sum\limits_{t=n+1}^{n+h} \left(y_t - \hat{y}_t \right)^2 }{ \frac{1}{n-1} \sum\limits_{t=2}^{n} \left( y_t - y_{t-1}\right)^2 }  }
\label{eqn:rmsse}
\end{equation}
where $y$ is the actual future value of the examined time series (for a specific aggregation level $l$) at point $t$ and $\hat{y}$ the predicted value,  $n$ the number of historical observations, and $h$ the $28$ day forecasting horizon. The choice of this metric is justified from the intermittency of forecasting data that involve sporadic unit sales with lots of zeros. 

The overall accuracy of each forecasting method at each aggregation level is computed by averaging the RMSSE scores across all the series of the dataset using appropriate price related weights. The measure, called weighted RMSSE (WRMSSE) by the organizers, is defined in Eq.~\ref{eqn:wrmsse}. 

\begin{equation}
\displaystyle
WRMSSE  = \sum\limits_{i=0}^{42,840} w_i \cdot {RMSSE}_i 
\label{eqn:wrmsse}
\end{equation}

\noindent where  $w_i$ is a weight assigned on the $i_{th}$ series. This weight is  computed based on the last $28$ observations of the training sample of the dataset, i.e., the cumulative actual dollar sales that each series displayed in that particular period (sum of units sold multiplied by their respective price)~
\cite{makridakis2022m5}. The weights $w_i$ are computed once and kept constant throughout the analysis.  

\subsection{Models}

Here we describe briefly the specific instances of GBM and neural network models used.
\subsubsection{LightGBM models}
LightGBM~\cite{ke2017lightgbm} is an open source Gradient Boosting Decision Tree (GBDT)~\cite{friedman2002stochastic,ye2009stochastic} implementation by Microsoft.
It uses a histogram-based algorithm to speed up the
training process and reduce memory. LightGBM models have proven to be very efficient in terms of speed and quality in many practical regression problems. For the Accuracy track, we trained two variations of LightGBM models:
\begin{enumerate}
    \item {\tt lgb\_cos}: Single lightGBM regression model for all available data (see Table~\ref{tab:lgb_cos_params})
    \item {\tt lgb\_nas}: A different lightGBM regression model for each store (10 models in total) (see Table~\ref{tab:lgb_nas_params})
\end{enumerate}

\begin{table}[!h]
\centering
\footnotesize
\begin{tabular}{|l|l|}
\hline
\textbf{parameter}          & \textbf{value} \\ \hline
boosting\_type            & gbdt         \\ \hline
 objective                 &  tweedie      \\ \hline
 tweedie\_variance\_power  & 1.1           \\ \hline
 subsample                 & 0.5           \\ \hline
 subsample\_freq           & 1             \\ \hline
 learning\_rate            & 0.03          \\ \hline
 num\_leaves             & 2047          \\ \hline
 min\_data\_in\_leaf       & 4095          \\ \hline
 feature\_fraction        & 0.5           \\ \hline
 max\_bin                & 100           \\ \hline
 n\_estimators            & 1300          \\ \hline
 boost\_from\_average     & False         \\ \hline
 verbose                  & -1            \\ \hline
 num\_threads            & 8              \\ \hline
\end{tabular}
\caption{Parameters for {\tt lgb\_cos} } \label{tab:lgb_cos_params}
\end{table}

\begin{table}[!htb]
\footnotesize
\begin{minipage}{.5\linewidth}
\caption{Parameters for {\tt lgb\_nas} }\label{tab:lgb_nas_params}
\centering
\begin{tabular}{|l|l|}
\hline
\textbf{parameters}         & \textbf{value} \\ \hline
boosting\_type           & gbdt       \\ \hline
objective              & tweedie     \\ \hline
tweedie\_variance\_power & 1.1          \\ \hline
subsample                & 0.6           \\ \hline
subsample\_freq          & 1             \\ \hline
learning\_rate           & 0.02          \\ \hline
num\_leaves             & 2**11-1       \\ \hline
min\_data\_in\_leaf      & 2**12-1       \\ \hline
feature\_fraction        & 0.6           \\ \hline
max\_bin                & 100          \\ \hline
n\_estimators            & see Table~\ref{tab:lgb_nas_estimators}             \\ \hline
boost\_from\_average     & False         \\ \hline
verbose                  & -1            \\ \hline
num\_threads            & 12             \\ \hline
\end{tabular}
\end{minipage}%
\begin{minipage}{.5\linewidth}
\caption{Number of tree estimators per store}\label{tab:lgb_nas_estimators}
\centering
\begin{tabular}{|l|l|}
\hline
\textbf{store\_id} & \textbf{n\_estimators} \\  \hline
CA\_1 & 700\\ \hline
CA\_2 & 1100\\ \hline
CA\_3 & 1600\\ \hline
CA\_4 & 1500\\ \hline
TX\_1 & 1000\\ \hline
TX\_2 & 1000\\ \hline
TX\_3 & 1000\\ \hline
WI\_1 & 1600\\ \hline
WI\_2 & 1500\\ \hline
WI\_3 & 1100\\  \hline
\end{tabular}
\end{minipage}

\end{table}

\begin{figure}[!h]
\centering
\includegraphics[scale=0.45]{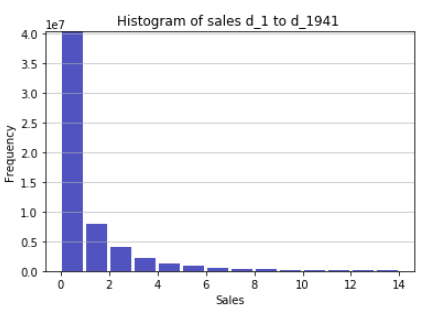}
\caption{Histogram of daily sales for Product/Store (Level 12) aggregation level} \label{fig:sales}
\end{figure}

The target output of each LightGBM models was the sales count of a specific product. Since the target quantity (sales count) is intermittent and has a lot of zeros ($ \approx 68\%$ of daily sales are zeros),  using the MSE loss function may lead to suboptimal solutions. For this reason, we implemented a special loss function that follows the Tweedie-Gaussian distribution~\cite{tweedie1957statistical}. Tweedie regression \cite{zhou2020tweedie} is designed to deal with right-skewed data where most of the target values are concentrated around zero. In Figure~\ref{fig:sales} we show the histogram of sales for all available training data; it is obviously right skewed with a lot of concentration around zero.
. 

The formula of the Tweedie loss function given a predefined parameter $p$ is shown in Eq.~\ref{eq:tweedie}. In our implementation we used the default value of $p=1.5$ which is a good balance between the two terms:
\begin{equation}
    \mathcal{L}(y, \hat{y}) = -\sum_{k} y^{(k)} \cdot \frac{{\left(\hat{y}^{(k)}\right)}^{1-p}}{1-p} + \frac{\left(\hat{y}^{(k)}\right)^{2-p}}{2-p}, \ \ \hat{y}^{(k)} = f(x^{(k)};\theta)
    \label{eq:tweedie}
\end{equation}

 where  $f(\cdot; \theta ) : \mathbf{R}^d \rightarrow \mathbf{R}$ is a regression model controlled by a set of parameters $\theta$ that maps a set of multidimensional features $x^{(k)} \in \mathbf{R}^d$ to a target value $y^{(k)}$ (daily sales count). $\hat{y}^{(k)}$ is the output of the regression model for input $x^{(k)}$.

\subsubsection{Neural Network models}
We implemented the following two classes of Neural Network models: 
\begin{itemize}
    \item  {{\tt keras\_nas} - Keras MLP models. }
    
    Keras~\cite{chollet2018deep} is a model-level library, providing high-level building blocks for developing neural network models. It has been adopted by Google to become the standard interface for  Tensorflow~\cite{abadi2016tensorflow}, its flagship Machine Learning library. Using the highly intuitive description of Keras building modules one can define a complex Neural Network architectures and experiment on training and inference.
    
    In total we trained $15$ slightly different Keras models and we averaged their predictions. These models were grouped in $3$ slightly different architecture groups, and within each group we kept the final $5$ weights during training.  This strategy helped reduce variance among predictions and stabilized the result, both in every validation split and in final training. The loss function chosen was mean squared error between the actual and predicted sale count. All model groups share the shame architecture depicted in Figure~\ref{fig:keras_model_basic} and are presented in detail at \url{https://github.com/IoannisNasios/M5_Uncertainty_3rd_place}.
    
    \begin{figure}[!h]
        \hspace{-3cm}
        \includegraphics[scale=0.06]{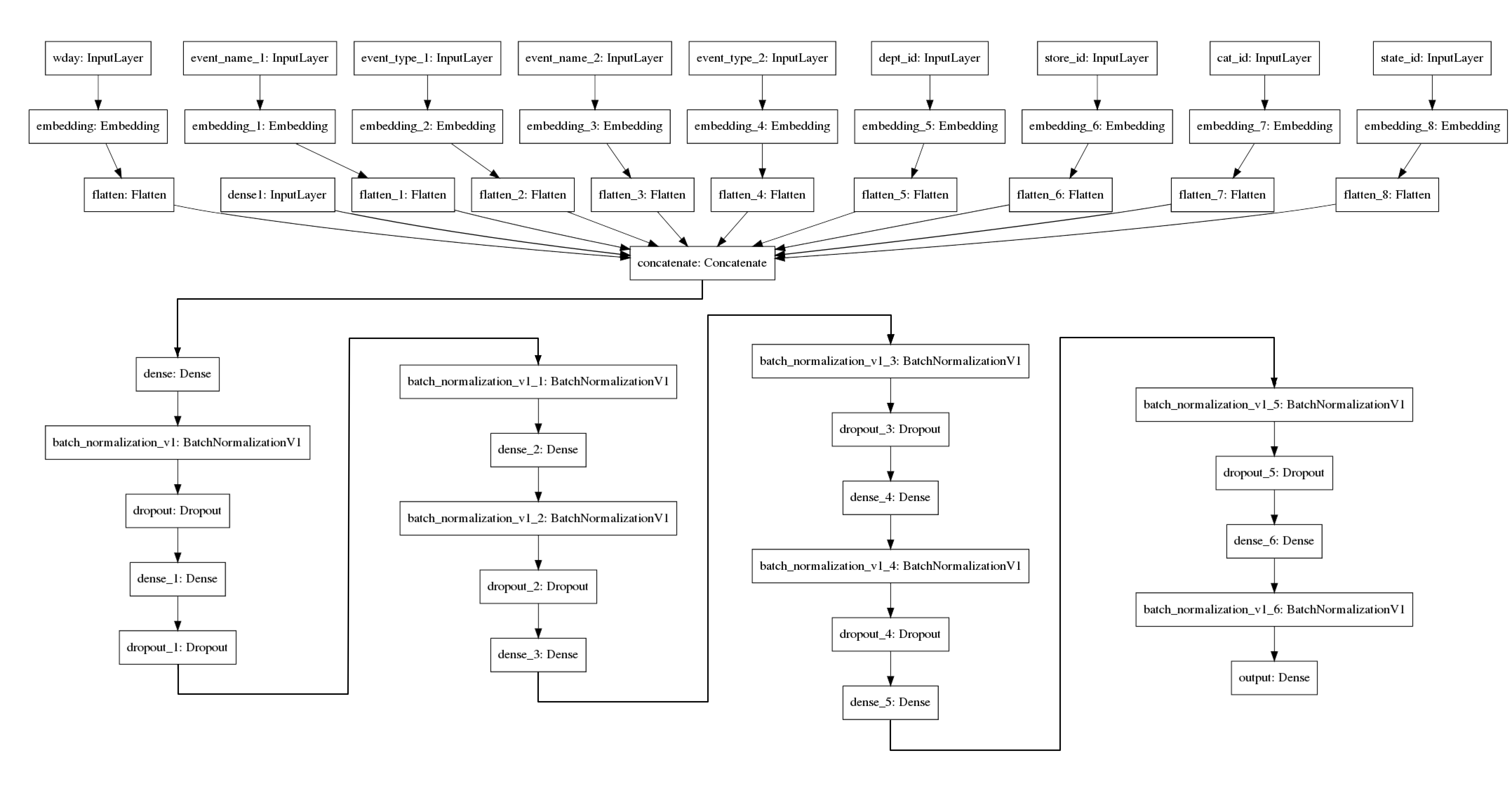}
        \caption{Basic Keras model architecture}
        \label{fig:keras_model_basic}
    \end{figure}

    \item  {{\tt fastai\_cos} - FastAI MLP models. }
    
    FastAI~\cite{howard2020fastai} is a Pytorch~\cite{paszke2019pytorch} based deep learning library which provides high-level components for many machine learning tasks. To tackle the regression task at hand, we incorporated the \emph{tabular} module that implements state of the art deep learning models for tabular data. One important thing about FastAI tabular module is again the use of embedding layers for categorical data. Similarly to the Keras case, using the embedding layer introduced good interaction for the categorical variables and leveraged deep learning’s inherent mechanism of automatic feature extraction.

    Our modelling approach includes implementing a special Tweedie loss function to be used during training.
    Roughly speaking, the FastAI tabular model is the Pytorch equivalent of Keras / Tensorflow {\tt keras\_nas} model with the difference of using a specially implemetned objective function. Empirical results on different regression contexts, mostly from Kaggle competitions, support this decision of using similar modelling methodology on complete different Deep Learning frameworks.
\end{itemize}

\subsection{Ensembling}
The final part of our modelling approach was to carefully blend the predictions of the diverge set of models. We divide our models in $4$ groups:
\begin{itemize}
\item {\tt lgb\_cos}: 1 LightGBM model, using all available data  
\item  {\tt lgb\_nas}:  10 LightGBM models (1 model per store), using all available data for every store.  
\item {\tt keras\_nas}: 3 Keras models using only last $17 \times 28$ days of data and simple averaging.  
\item {\tt fastai\_cos}: 1 FastAI model, using all available data  
\end{itemize}

All these 4 model groups were finetuned  to perform best on the average of the three validation splits described in Section~\ref{sec:validation} above. After fine-tuning, and using the best parameters,  all 4 model groups were retrained using the information available until $d_{1941}$ and then used to produce forecasts for days $d_{1942}$ until $d_{1969}$.

The individual predictions were blended using geometric averaging shown in Eq.~\ref{eq:blend}.
 \begin{equation}
       \textrm{blend} = \begin{cases} 
       \left (\textrm{lgb\_nas}^{3.5}  \cdot 
        \textrm{lgb\_cos}^1  \cdot 
       \textrm{keras\_nas}^1 \cdot 
       \textrm{fastai\_cos}^{0.5} \right)^{
       \frac{1}{6}} 
        & \textrm{for days 1-27} \\ \\
        \left (\textrm{lgb\_nas}^{3}    \cdot 
        \textrm{lgb\_cos}^{0.5}  \cdot 
       %\textrm{keras\_nas}^0 \cdot 
       \textrm{fastai\_cos}^{1.5} \right)^{
       \frac{1}{5}} & \textrm{for day 28}
   \end{cases}
   \label{eq:blend}
 \end{equation}

By reviewing several  predictions by the {\tt keras\_nas} group of models, we noticed an unexpected large peak on the last day of the private set (day $d_{1969}$). After confirming this behavior to be dominant in almost all cases, we decided to exclude that group’s prediction for the last day. This is the reasoning for not using {\tt keras\_nas} predictions on the lower branch of Eq~\ref{eq:blend}. In Table~\ref{tab:blend_weights} we present the weights for each model component as well as the validation WRMSSE score for each validation split. 

\begin{table}[!ht]
\footnotesize
\begin{tabular}{|l|c|l|c|l|c|l|c|l|c|l|}
\hline
\textbf{}              & \multicolumn{2}{c|}{{\tt lgb\_nas}} & \multicolumn{2}{c|}{ \tt lgb\_cos} & \multicolumn{2}{c|}{\tt keras\_nas} & \multicolumn{2}{c|}{\tt fastai\_cos} & \multicolumn{2}{c|}{Ensemble}             \\ \hline
weights $d_{1}-d_{27}$ & \multicolumn{2}{c|}{3.5}               & \multicolumn{2}{c|}{1.0}      & \multicolumn{2}{c|}{1.0}        & \multicolumn{2}{c|}{0.5}         & \multicolumn{2}{c|}{\multirow{2}{*}{Eq.\ref{eq:blend}}} \\ \cline{1-9}
weight $d_{28}$        & \multicolumn{2}{c|}{3.0}               & \multicolumn{2}{c|}{0.5}      & \multicolumn{2}{c|}{0.0}        & \multicolumn{2}{c|}{1.5}         & \multicolumn{2}{c|}{}                  \\ \hline

\end{tabular}
\label{tab:blend_weights} \caption{Final weights selected for blending}
\end{table}

\begin{table}[!ht]
\footnotesize
\begin{tabular}{|l|c|l|c|l|c|l|c|l|c|l|}
\hline
\textbf{}              & \multicolumn{2}{c|}{{\tt lgb\_nas}} & \multicolumn{2}{c|}{ \tt lgb\_cos} & \multicolumn{2}{c|}{\tt keras\_nas} & \multicolumn{2}{c|}{\tt fastai\_cos} & \multicolumn{2}{c|}{Ensemble}             \\ \hline

Val. split 1             & \multicolumn{2}{c|}{0.474}             & \multicolumn{2}{c|}{0.470}    & \multicolumn{2}{c|}{0.715}      & \multicolumn{2}{c|}{0.687}       & \multicolumn{2}{c|}{0.531}             \\ \hline
Val. split 2             & \multicolumn{2}{c|}{0.641}             & \multicolumn{2}{c|}{0.671}    & \multicolumn{2}{c|}{0.577}      & \multicolumn{2}{c|}{0.631}       & \multicolumn{2}{c|}{0.519}             \\ \hline
Val. split 3             & \multicolumn{2}{c|}{0.652}             & \multicolumn{2}{c|}{0.661}    & \multicolumn{2}{c|}{0.746}      & \multicolumn{2}{c|}{0.681}       & \multicolumn{2}{c|}{0.598}             \\ \hline
Mean / std             & 0.589              & 0.08              & 0.602          & 0.09         & 0.679           & 0.05          & 0.667           & 0.02           & 0.549              & 0.03              \\ \hline
\end{tabular}
\label{tab:blend_scores} \caption{Competition scores for each component}
\end{table}
    
\subsection{Post processing}
We tried several post processing smoothing techniques to improve forecasting accuracy. In the final solution we employed a simple exponential smoothing ($\alpha = 0.96$) per product and store id (Level 12 - $30,490$) that lead to substantial improvement in both validation sets and the final  evaluation (private leaderboard). 

We notice that, although this post-processing should be performed for both competition tracks, we were able to only use it for the Uncertainty track as it was a last minute finding (2 hours before competitions closing); as it turned out, had we applied it also for the Accuracy track submissions we would have ended 3 positions higher in the Accuracy track leaderboard.

\section{Probabilistic Forecast Methodology}

The performance measure selected for this competition track was the Scaled Pinball Loss (SPL) function. The measure is calculated for the 28 days horizon for  each series and quantile, as shown in  Eq.~\ref{eqn:spl}:
\begin{equation}
\displaystyle
SPL =  \frac{1}{h} \frac{ \sum_{t=n+1}^{n+h} \left(Y_t - Q_t(u)\right) u \textbf{1}\{Q_t(u) \leq Y_t\} + \left(Q_t(u)-Y_t\right)(1-u)\textbf{1}\{Q_t(u) > Y_t \}  } { \frac{1}{n-1} \sum_{t=2}^2 \left |Y_t - Y_{t-1} \right| }
\label{eqn:spl}
\end{equation}
where $Y_t$ is the actual future value of the examined time series at point $t$, $Q_t(u)$ the generated forecast for quantile $u$, $h$ the forecasting horizon, $n$ is the number of historical observations, and $\textbf{1}$ the indicator function (being 1 if Y is within the postulated interval and 0 otherwise). The  values $u$ were set to $u_1=0.005$, $u_2=0.025$, $u_3=0.165$, $u_4=0.25$, $u_5=0.5$, $u_6=0.75$, $u_7=0.835$, $u_8=0.975$, and $u_9=0.995$, so that they correspond to the requested median,  $50\%$, $67\%$, $95\%$, and $99\%$ prediction intervals.

After estimating the SPL for all time series and  all the requested quantiles of the competition, the Uncertainty track competition entries were be ranked using the Weighted SPL (WSPL) shown in Eq.~\ref{eqn:wspl}: 

\begin{equation}
\displaystyle
WSPL =  \sum_{i=1}^{42,840} w_i * \frac{1}{9} \sum_{j=1}^9 SPL(u_j)
\label{eqn:wspl}
\end{equation}

\noindent where weights $w_i$ were same described in Section~\ref{sec:accuracy}.

\subsection{Quantile estimation via optimization}
\label{sec:quantestim}
Using our best point forecast as median, we proceed on optimizing WSPL objective function on validation split 1 (last 28 days) and calculate the factors in Table~\ref{tab:quantiles}. Due to time restrictions of the competition we could not extend our analysis to cover all three validation splits. These factors were used to multiply median solution (quantile $u=0.50$) and produce the remaining upper and lower quantiles. We assumed symmetric distributions on levels 1-9 and skewed distributions on levels 10-12.  Furthermore, due to right-skewness of our sales data (zero-bounded on the left) for every level, on last quantile ($99.5\%$) distributions  proposed factor was multiplied by either $1.02$ or $1.03$. These factors were determined so as to minimize WSPL on validation split 1.

\begin{table}[!ht]
\begin{tiny}
\begin{tabular}{p{0.25cm}p{1.3cm}p{0.35cm}ccccccccc}
\hline
\textbf{Level} & \textbf{Aggr} & \# & \textbf{0.005} & \textbf{0.025} & \textbf{0.165} & \textbf{0.25} & \textbf{0.5} & \textbf{0.75} & \textbf{0.835} & \textbf{0.975} & \textbf{0.995} \\ \hline
1 & Total                              & 1        & 0.890          & 0.922          & 0.963          & 0.973         & 1.000        & 1.027         & 1.037          & 1.078          & 1.143          \\
2 & State                              & 3        & 0.869          & 0.907          & 0.956          & 0.969         & 1.000        & 1.031         & 1.043          & 1.093         & 1.166          \\
3 & Store                              & 10       & 0.848          & 0.893          & 0.950          & 0.964         & 1.000        & 1.036         & 1.049          & 1.107          & 1.186          \\
4 & Category                           & 3        & 0.869          & 0.907          & 0.951          & 0.969         & 1.000        & 1.031         & 1.043          & 1.093          & 1.166          \\
5 & Dept.                             & 7        & 0.827          & 0.878          & 0.943          & 0.960         & 1.000        & 1.040         & 1.057          & 1.123          & 1.209          \\
6 &  State/Cat.                     & 9        & 0.827          & 0.878          & 0.943          & 0.960        & 1.000        & 1.040         & 1.057          & 1.123          & 1.209          \\
7 & State/Dept.                    & 21       & 0.787          & 0.850          & 0.930          & 0.951         & 1.000        & 1.048         & 1.070          & 1.150          & 1.251          \\
8 & Store/Cat.                     & 30       & 0.767          & 0.835          & 0.924          & 0.947         & 1.000        & 1.053         & 1.076          & 1.166          & 1.272     \\
9 & Store/Dept.                    & 70       & 0.707          & 0.793          & 0.905          & 0.934         & 1.000        & 1.066         & 1.095          & 1.208          & 1.335          \\
10 & Product                            & 3.049    & 0.249          & 0.416          & 0.707          & 0.795         & 1.000        & 1.218         & 1.323          & 1.720          & 2.041          \\
11 & Product/State                      & 9.147    & 0.111          & 0.254          & 0.590          & 0.708         & 1.000        & 1.336         & 1.504          & 2.158          & 2.662          \\
12 & Product/Store                      & 30.490   & 0.005          & 0.055          & 0.295         & 0.446        & 1.000        & 1.884         & 2.328          & 3.548          & 4.066          \\
\hline
\end{tabular}
\end{tiny}
\caption{Quantile factors for all levels}\label{tab:quantiles}
\end{table}

\subsection{Quantile correction via statistical information}
Level 12 (the only non-aggregated one) was the most difficult for accurate estimations, and the previously calculated factors could be further improved. Statistical information on past days played a major role for this level.
For every product/store id, eight statistical sales quantiles (four intervals excluding median) were calculated over the last $13 \times 28$ days ($1$ year) and over the last $28$ days. These quantiles were first averaged and then used to correct the quantile estimation of Section~\ref{sec:quantestim} for the corresponding level. For the same reason we calculated {\emph weekly sales quantiles} over the last $13 \times 28$ days and $3 \times 28$ days.  The final formula for estimating level $12$ quantiles, which led to minimum WSPL score for validation split 1, is given below:
\begin{footnotesize}
\begin{eqnarray*}
\displaystyle
  \textrm{level 12} &=& 0.2\cdot\textrm{quantile estimation}\\
                    &+& 0.7 \cdot \left(  \frac{\textrm{statistical correction}_{13 \times 28 \textrm{ days}}  +                1.75\cdot\textrm{statistical correction}_{28 \textrm{ days}} }{2.75}\right) \\
                    &+& 0.1 \cdot  \left(  \frac{\textrm{weekly statistical correction}_{13 \times 28 \textrm{ days}} + \textrm{weekly statistical correction}_{3 \times 28 \textrm{ days}}}{2} \right) %\textrm{weekly statistical correction}
\end{eqnarray*}
\end{footnotesize}

Level $11$ was also corrected in a similar manner and the final formula is given below:
\begin{footnotesize}
\begin{eqnarray*}
  \textrm{level 11} &=& 0.91 \cdot \textrm{quantile estimation} \\
                              &+& 0.09 \cdot  \left(  \frac{\textrm{statistical correction}_{13 \times 28 \textrm{ days}}  +                1.75\cdot\textrm{statistical correction}_{28 \textrm{ days}} }{2.75}\right)  
\end{eqnarray*}
\end{footnotesize}

No corrections were applied for levels other than $12$ and $11$, so the respective quantile factors for levels 1-10 remain as shown in Table~\ref{tab:quantiles}.

\begin{figure}[!h]
\centering
\hspace*{-1cm}\includegraphics[scale=0.45]{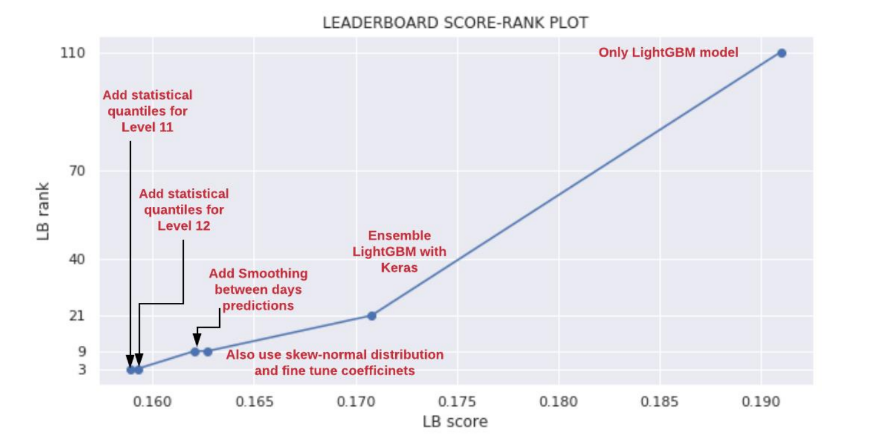}
\caption{Leaderboard ranking vs. score}\label{fig:exelixi}
\end{figure}

The evolution of our submission attempts (SPL score / ranking) for the Uncertainty track of the competition is shown in Fig.~\ref{fig:exelixi}. This scatter plot highlights: 
\begin{enumerate}[a)]
    \item  the importance of ensembling diverse models, as our rank increased by almost 90 places only by ensembling,
    \item the contribution of statistical correction of levels 11 and 12 that led us to a winning placement.
\end{enumerate}

\begin{figure}[!h]
\centering
\hspace*{-1cm}\includegraphics[scale=0.525]{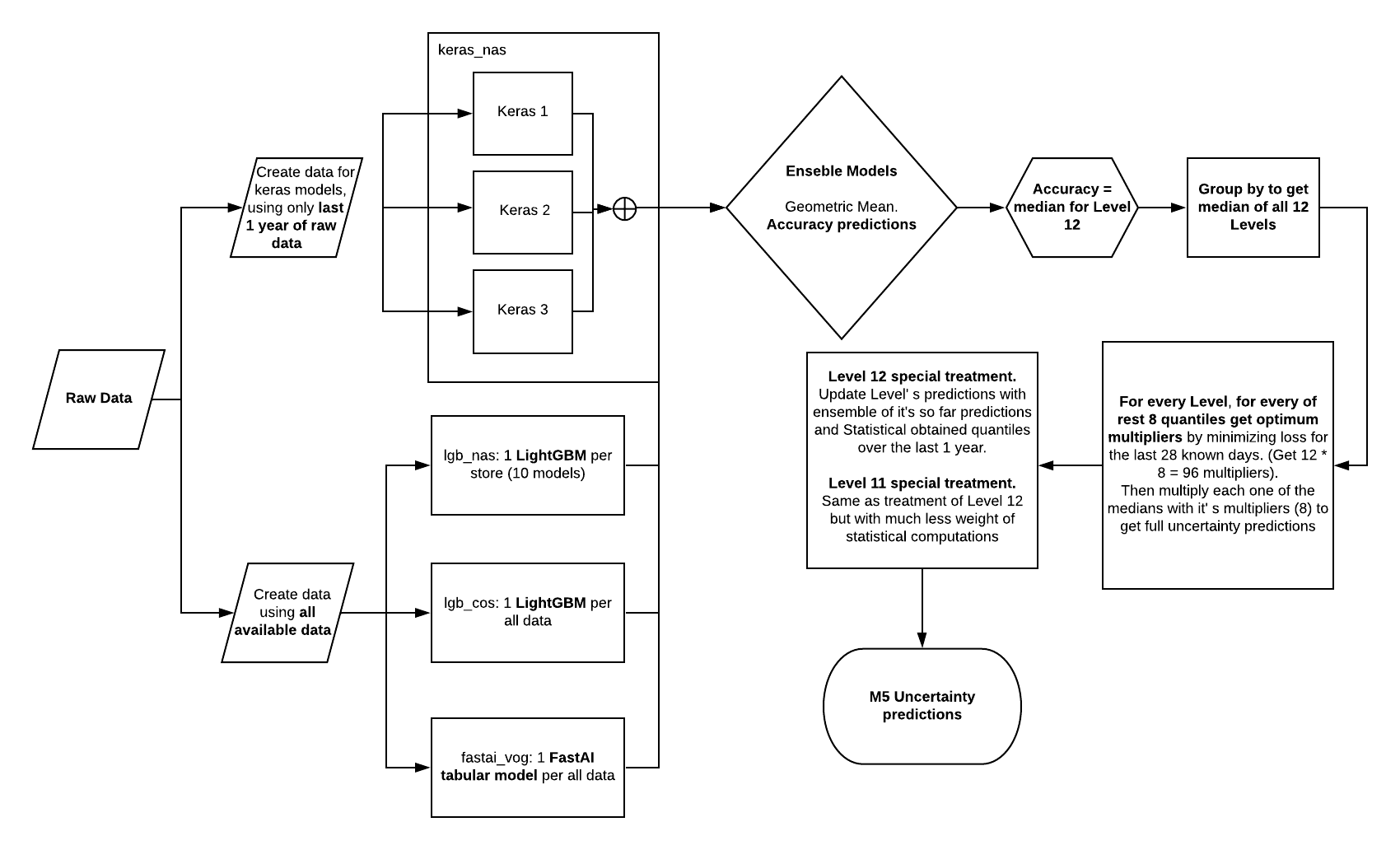}
\caption{Flowchart of probabilistic forecasting pipeline}\label{fig:diagram}
\end{figure}

The overall procedure of ensembling, optimizing, and correcting for the probabilistic forecasting is depicted in Fig.~\ref{fig:diagram}.

\section{Discussion}
We have presented a Machine Learning solution for point and probabilistic forecasting of hierarchical time series that represent daily unit sales of retail products. The methodology involves two state-of-the-art Machine Learning approaches, namely Gradient Boosting Trees and Neural Networks, tuned and combined using carefully selected training and validation sets. The proposed methodology was applied successfully on the recent M5 Competition, yielding a prize position placement in the Uncertainty track. 

\subsection{Point Forecasting score breakdown}
\begin{figure}[!h]
    \centering
    \includegraphics[scale=0.45]{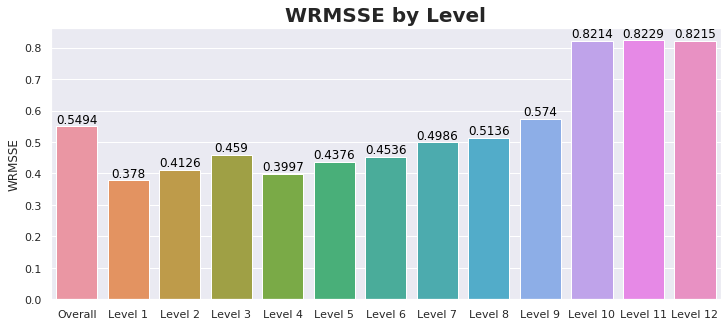}
    \caption{Loss per Level for predictions of validation split 1 }
    \label{fig:score_per_level}
\end{figure}

In order to get a deeper insight on the point forecasting task, we broke down the WRMSSE calculation of Eq.~\ref{eqn:wrmsse} on each hierarchical level. The resulting per-level WRMSSE for validation split 1 is shown in Fig.~\ref{fig:score_per_level}.
It is obvious that the performance varies significantly among different aggregation levels, with levels $10$, $11$ and $12$ being the harder to predict. An interesting observation is that, although Level 1 is simply the aggregation of all Level $12$ predictions, the Level $1$  loss is less than half the magnitude of the Level $12$ loss. It would  seem that aggregation \emph{cancels out} the poor predictive capability on Level $12$. This is stressed out even more by the fact that, even though mixture of traditional forecasting techniques (ARIMA, Exponential smoothing) achieved lower error on levels 10, 11, 12, than the one shown in Figure~\ref{fig:score_per_level}, the overall score was really poor. 
 
The mean validation WRMSSE score using the final blend of Eq.~\ref{eq:blend} is $0.549$, while our final submission score was $0.552$ -  only $0.003$ units apart. This close tracking of the unseen test data error by the validation error is generally sought after by ML practitioners in real world problems, and it serves as an extra indication for the soundness of our methodology. 

\subsection{Probabilistic Forecasting Factors visualized}
\begin{figure}[!h]
    \centering
    \includegraphics[scale=0.35]{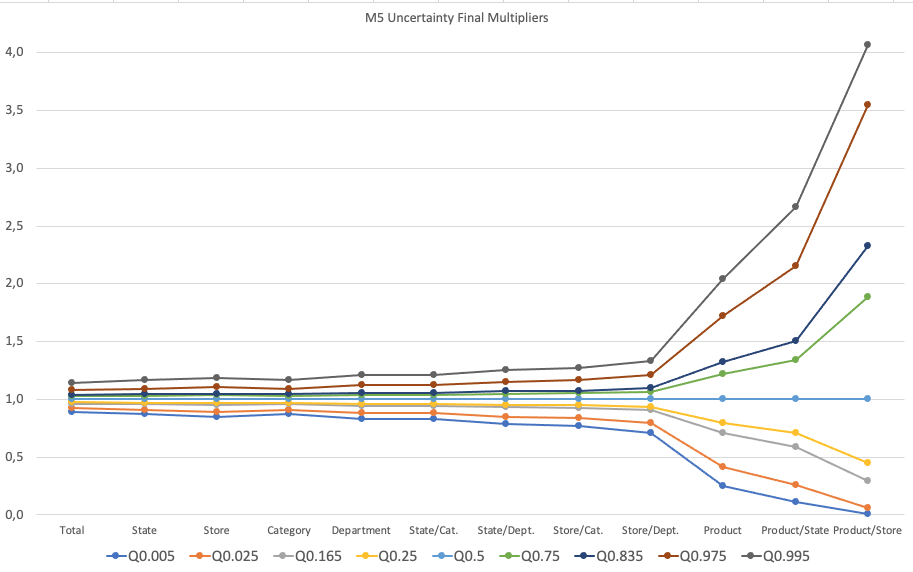}
    \caption{Quantile factors for increasing levels of aggregation  }
    \label{fig:uncertainty_multipliers}
\end{figure}

In Fig.~\ref{fig:uncertainty_multipliers} we present a graphical plot of the factors sorted by increasing level. We notice an unsymmetrical widening of the calculated factors as the number of series of the corresponding level increases. This delta-like shape is correlated to the increasing WRMSSE error on levels 10-12 shown in Fig.~\ref{fig:score_per_level}.

\subsection{Takeaways}
It is worth mentioning that our point and probabilistic forecasting results are highly correlated since we use the point forecasting as the starting point for our probabilistic analysis. It is expected that, starting from better point forecasts, will result to better probabilistic forecasts as well. 

Model diversity and selected training/validation split was crucial for the overall performance and choices eliminated any need for external adjustments on the final prediction in both tracks. This is also highlighted in M5 Competition Summary~\cite{makridakis2022m5}. After competition ended we found out that by using the magic multiplier $0.97$ we would end up first in Accuracy Track.

Finally, although the M5 Competition dataset was hierarchical in nature, we have not used this information explicitly via a reconciliation procedure as described in~\cite{hyndman2014optimally, hyndman2018forecasting}. Hierarchy was only implicitly considered via the nature of WRMSSE objective function.    

\subsection{Extensions}
The competition data were carefully curated and clean providing a rich variety of potential features. There were also a multitude of ML regression methods that could be tested. However we feel that the most crucial decision was not the selection of features or ML models but the selection of a representative validation sets. Although the three validation splits described in Section~\ref{sec:validation} were enough to stabilize our point forecasting results we believe that using more validation splits could be beneficial and would even eliminate the need of \emph{magic} external multipliers ($0.97$) to reach the top place. 

One  obvious extension to the probabilistic forecast methodology was to use a weighted averaged on three validation splits instead of validation  split  1  (last  28  days).

\bibliographystyle{elsarticle-num-names}
\bibliography{sample.bib}

\newpage
\appendix

\section{Complete feature list}
\begin{footnotesize}
\setlength\LTleft{-2cm}

\begin{longtable}{|p{2.5cm}|p{3.2cm}|l|p{5cm}|c|c|c|}
\caption{M5 competition features}
\label{tab:my-table}\\
\hline
{\bf feature category } & {\bf feature name}   & {\bf type}    & {\bf meaning}    &  {\bf lightGBM} & {\bf keras} & {\bf fastai}   \\ \hline
\endfirsthead
\multicolumn{5}{c}%
{{\bfseries Table \thetable\ continued from previous page}} \\
\hline
feature category & feature name & type & meaning  &  lightGBM & keras & fastai   \\ \hline
\endhead
\multirow{2}{2.5cm}{non trainable features} & id  & cat     & 30490 unique ids that correspond to combination of item\_id/store\_id                                                                                                                                                                                                                                                                   &   &  &                                                                                                                                                                                                    \\ \cline{2-7}
                                                         & d                                 & integer & the increasing number of days 1-1941 &  &  & \\ \hline
\multirow{5}{2.5cm}{id based categorical features}           & item\_id                        & cat     & 3049 distinct item ids                                                                                                                                                                                                                                                                                                                  & * &  & *  \\\cline{2-7}
                                                         & dept\_id                       & cat     & 7 distinct departments     & * & * & *  \\ \cline{2-7}
                                                         & cat\_id                        & cat     & 3 product categories             & * & * &  *    \\ \cline{2-7}
                                                         & store\_id                      & cat     & 10 different stores                 & * & * &  * \\ \cline{2-7}
                                                         & state\_id                      & cat     & 3 different states                   &   & * & \\ \hline

\multirow{13}{2.5cm}{price related features}                 & release                        & integer & week when the first price was set of the product                                                                                                                                                                                                                                                                                         & * &  & *                                                                                                                                                                                                                                 \\ \cline{2-7} 
                                                         & sell\_price                    & float   & current price of the product                                                                                                                                                                                                                                                                                                            & * & * &  *                                                                                                                                                                                                                               \\ \cline{2-7} 
                                                                                                                  & sell\_price\_rel\_diff                    & float   & relative difference between two consecutive price changes                                                                                                                                                                                                                                                                                                    &  & * &                                                                                                                                                                                                                                 \\ \cline{2-7} 

                                                         & price\_max                     & float   & maximum price this product ever reached                                                                                                                                                                                                                                                                                                 & * &  & *                                                                                                                                                                                                                                \\ \cline{2-7} 
                                                         & price\_min                     & float   & minimum price this product ever reached                                                                                                                                                                                                                                                                                                 & * &  &  *                                                                                                                                                                                                                               \\ \cline{2-7} 
                                                         & price\_std                     & float   & standard deviation of the historical prices of the product                                                                                                                                                                                                                                                                              & * &  &  *                                                                                                                                                                                                                            \\ \cline{2-7} 
                                                         & price\_mean                    & float   & average value of the historical prices of the product                                                                                                                                                                                                                                                                                   & * &  &  *                                                                                                                                                                                                                              \\ \cline{2-7} 
                                                         & price\_norm                    & float   & normalized price (price/price\_max) for the day {[}0 1{]}                                                                                                                                                                                                                                                                              & * &  &  *                                                                                                                                                                                                                            \\ \cline{2-7} 
                                                         & price\_nunique                 & integer & number of unique prices for this product                                                                                                                                                                                                                                                                                                & * &  &  *                                                                                                                                                                                                                                  \\ \cline{2-7} 
                                                         & item\_nunique                  & integer & number of unique products that reached the same price in a specific store.                                                                                                                                                                                                                                                              & * &  &  *                                                                                                                                                                                                                                \\ \cline{2-7} 
                                                         & price\_momentum                & float   & ratio of the previous to the current price of the product                                                                                                                                                                                                                                                                               & * &  &  *                                                                                                                                                                                                                                      \\ \cline{2-7} 
                                                         & price\_momentum\_m             & float   & mean price of the product the last month                                                                                                                                                                                                                                                                                                & * &  &   *                                                                                                                                                                                                                                     \\ \cline{2-7} 
                                                         & price\_momentum\_y             & float   & mean price of the product the last year                                                                                                                                                                                                                                                                                                 & * &  &   *                                                                                                                                                                                                                                    \\ \hline
\multirow{15}{2.5cm}{calendar related features}              & event\_name\_1                 & cat     & If the date includes an event the name of this event                                                                      & * &  &   * \\ \cline{2-7} 
                                                         & event\_type\_1                 & cat     & If the date includes an event the type of this event                & * & * &   *\\ \cline{2-7} 
                                                         & event\_name\_2                 & cat     & If the date includes a second event the name of this event             &  * & * &    *   \\ \cline{2-7} 
                                                         & event\_type\_2                 & cat     & If the date includes a second event the type of this event.          &  * & * & *\\ \cline{2-7} 
                                                         & snap\_CA                       & binary  & SNAP activities that serve as promotions in California on a specific date. &  * & * & * \\ \cline{2-7} 
                                                         & snap\_TX                       & binary  & SNAP activities that serve as promotions in Texas on a specific date.      &  * &  &  *   \\ \cline{2-7} 
                                                         & snap\_WI                       & binary  & SNAP activities that serve as promotions in Wiskonskin on a specific date.   &  * &  & *   \\ \cline{2-7} 
                                                         & log\_d                          & float & $log( 1 + d)$ where $d$ the increasing number of days &   & * &    \\ \cline{2-7} 
                                                         & tm\_d                          & integer & day of the month for a specific date &  * &  &  *  \\ \cline{2-7} 
                                                         & tm\_w                          & integer & week of the year for a specific date  & * &  & *   \\ \cline{2-7} 
                                                         & tm\_m                          & integer & month of a year for a specific date  & * &  & *   \\ \cline{2-7} 
                                                         & tm\_y                          & integer & year of the specific date (5 years total) &  * &  &  *  \\ \cline{2-7} 
                                                         & tm\_wm                         & integer & number of week within a month    & * &  &  * \\ \cline{2-7} 
                                                         & tm\_dw                         & integer & number of the day within a week         &  * & * & * \\ \cline{2-7} 
                                                         & tm\_w\_end                     & binary  & is weekend or not                     &  * &  &  \\ \hline
\multirow{12}{2.5cm}{target encoding features}               & enc\_cat\_id\_mean             & float   & average value of sales for each category id                                                                                                                                                                                                                                                                                             &  * &  &                                                                                                                                                                                                            \\ \cline{2-7} 
                                                         & enc\_cat\_id\_std              & float   & std value of sales for each category id                                                                                                                                                                                                                                                                                                 &  * &  &                                                                                                                                                                                                              \\ \cline{2-7} 
                                                         & enc\_dept\_id\_mean            & float   & average value of sales for each department id                                                                                                                                                                                                                                                                                           &  * &  &                                                                                                                                                                            \\ \cline{2-7} 
                                                         & enc\_dept\_id\_std             & float   & std value of sales for each department id                                                                                                                                                                                                                                                                                               &  * &  &                                                                                                                                                                                 \\ \cline{2-7} 
                                                         & enc\_item\_id\_mean            & float   & average value of sales for each item id                                                                                                                                                                                                                                                                                                 &  * &  &                                                                                                                                                                                                               \\ \cline{2-7} 
                                                         & enc\_item\_id\_std             & float   & std value of sales for each department ids                                                                                                                                                                                                                                                                                              &  * &  &                                                                                                                                                                                                                \\ \cline{2-7} 
                                                         & enc\_item\_id\_state\_id\_mean & float   & average value of sales for each combination of item id and state id                                                                                                                                                                                                                                                                     &  * &  &                                                                                                                                                                                                              \\ \cline{2-7} 
                                                         & enc\_item\_id\_state\_id\_std  & float   & std value of sales for each combination of item id and state id                                                                                                                                                                                                                                                                         &  * &  &                                                                                                                                                                                                              \\ \cline{2-7} 
                                                         & enc\_state\_id\_dept\_id\_mean & float   & average value of sales for each combination of item id and department id                                                                                                                                                                                                                                                                & * &  & *                                                                                                                                                                                                                \\ \cline{2-7} 
                                                         & enc\_state\_id\_dept\_id\_std  & float   & std value of sales for each combination of item id and department id                                                                                                                                                                                                                                                                    &  * &  & *                                                                                                                                                                                                              \\ \cline{2-7} 
                                                         & enc\_state\_id\_cat\_id\_mean  & float   & average value of sales for each combination of item id and category id                                                                                                                                                                                                                                                                  &  * &  &                                                                                                                                                            \\ \cline{2-7} 
                                                         & enc\_state\_id\_cat\_id\_std   & float   & std value of sales for each combination of item id and category id                                                                                                                                                                                                                                                                      &  * &  &                                                                                                                                                                     \\ \hline
\multirow{15}{2.5cm}{lag features}                           & sales\_lag\_28                 & integer & \multirow{15}{5cm}{the sales value for each id and each date 28 29 30 ... 42 days ago (span 2 weeks)}                                                                                                                                                                                                                                 & \multirow{15}{*}{} *  & *&*                                                                                                                                                                                                                 \\ \cline{2-3} \cline{5-7} 
                                                         & sales\_lag\_29                 & integer &                                                                 &   & &\\ \cline{2-3}\cline{5-7} 
                                                         & sales\_lag\_30                 & integer &  & *& &*   \\ \cline{2-3}\cline{5-7} 
                                                         & sales\_lag\_31                 & integer &   & *&  &* \\ \cline{2-3}\cline{5-7} 
                                                         & sales\_lag\_32                 & integer &    & *&  &*   \\ \cline{2-3}\cline{5-7} 
                                                         & sales\_lag\_33                 & integer &   & *&  &*  \\ \cline{2-3}\cline{5-7} 
                                                         & sales\_lag\_34                 & integer &  & *& &*   \\ \cline{2-3}\cline{5-7} 
                                                         & sales\_lag\_35                 & integer &   & *&  &*    \\ \cline{2-3}\cline{5-7} 
                                                         & sales\_lag\_36                 & integer &   & *& &* \\ \cline{2-3}\cline{5-7} 
                                                         & sales\_lag\_37                 & integer & & *&  &*  \\ \cline{2-3}\cline{5-7} 
                                                         & sales\_lag\_38                 & integer &  & *&  &*   \\ \cline{2-3}\cline{5-7} 
                                                         & sales\_lag\_39                 & integer &   & *&  &*  \\ \cline{2-3}\cline{5-7} 
                                                         & sales\_lag\_40                 & integer &  & *& &*  \\ \cline{2-3}\cline{5-7} 
                                                         & sales\_lag\_41                 & integer &  & *&   &*   \\ \cline{2-3} \cline{5-7} 
                                                         & sales\_lag\_42                 & integer &     & *&   &*    \\ \hline
\multirow{12}{2.5cm}{rolling statistics 28 days ago}         & rolling\_mean\_7               & float   & \multirow{10}{5.0cm}{rolling means and averages for the sales of a specific product that end 28 days prior current date}    & * & * & * 
%\multirow{10}{3.0cm}{for example rolling\_mean\_7 for a specific id on a specific date d is  calculated using the sales of the specific id on dates d-29 d-30 d-31 d-32 d-33 d-34 d-35} 
\\ \cline{2-3} \cline{5-7} 
                                                & rolling\_median\_7                & float   &  &  & * &    \\ \cline{2-3}\cline{5-7} 

                                                         & rolling\_std\_7                & float   &  & * &  & *   \\ \cline{2-3}\cline{5-7} 
                                                         & rolling\_mean\_14              & float   &  & * & & *      \\ \cline{2-3}\cline{5-7} 
                                                         & rolling\_std\_14               & float   &     & * & &*    \\ \cline{2-3}\cline{5-7} 
                                                         & rolling\_mean\_30              & float   &   &  * & * & *   \\ \cline{2-3}\cline{5-7} 
                                                     & rolling\_median\_30              & float   &   &   & * &    \\ \cline{2-3}\cline{5-7} 

                                                         & rolling\_std\_30               & float   &   &  * & & *  \\ \cline{2-3}\cline{5-7} 
                                                         & rolling\_mean\_60              & float   &    & * & & *  \\ \cline{2-3}\cline{5-7} 
                                                         & rolling\_std\_60               & float   &     & * & & *  \\ \cline{2-3} \cline{5-7} 
                                                         & rolling\_mean\_180             & float   &    & * & & *  \\ \cline{2-3} \cline{5-7} 
                                                         & rolling\_std\_180              & float   &       &  * & & *   \\ \hline
\multirow{12}{2.5cm}{rolling means within the 28 day window} & rolling\_mean\_tmp\_1\_7       & float   & \multirow{12}{5.0cm}{rolling means and averages for the sales of a specific product within 28 days prior current date.  While we can train our models with these features we need to be extra careful when predicting 28  days ahead because these features need to be calculated recursively.} &

%\multirow{12}{3.0cm}{for example rolling\_mean\_tmp\_1\_7 for a specific id on a specific date d is  calculated using the sales of the specific id on dates  d-2 d-3 d-4 d5 d-6 d-7 d-8}
* & & 
\\ \cline{2-3} \cline{5-7}
                                                         & rolling\_mean\_tmp\_1\_14      & float   & & * & & *  \\ \cline{2-3}\cline{5-7}
                                                         & rolling\_mean\_tmp\_1\_30      & float   &  & * & & *  \\ \cline{2-3}\cline{5-7}
                                                         & rolling\_mean\_tmp\_1\_60      & float   &   & * & & *   \\ \cline{2-3}\cline{5-7}
                                                         & rolling\_mean\_tmp\_7\_7       & float   &  & * & & *  \\ \cline{2-3}\cline{5-7}
                                                         & rolling\_mean\_tmp\_7\_14      & float   &  & * & & *   \\ \cline{2-3}\cline{5-7}
                                                         & rolling\_mean\_tmp\_7\_30      & float   & &  * & & * \\ \cline{2-3}\cline{5-7}
                                                         & rolling\_mean\_tmp\_7\_60      & float   &  & * & &  *  \\ \cline{2-3}\cline{5-7}
                                                         & rolling\_mean\_tmp\_14\_7      & float   &  & * & & *  \\ \cline{2-3}\cline{5-7}
                                                         & rolling\_mean\_tmp\_14\_14     & float   &  & * & & * \\ \cline{2-3}\cline{5-7}
                                                         & rolling\_mean\_tmp\_14\_30     & float   &  & * & & *  \\ \cline{2-3}\cline{5-7}
                                                         & rolling\_mean\_tmp\_14\_60{]}   & float   &   &  * & & * \\ \hline
\end{longtable}
\end{footnotesize} 

\end{document}